\documentclass[12pt,letterpaper]{article}
\usepackage{setspace} 
\usepackage[left=1in,right=1in,top=1in,bottom=1in]{geometry}
\usepackage{algorithm,algorithmic} 
\usepackage{xurl}
\usepackage{abstract}
\usepackage{cite}

\usepackage{lipsum}

\usepackage{amsthm,bbm}
\usepackage{ragged2e}
\usepackage{booktabs, multirow, tabularx, lmodern, babel, multirow, adjustbox}
    \newcolumntype{L}{>{\raggedright\arraybackslash}X}  

\usepackage{graphicx}
\usepackage{titlesec}
\usepackage{threeparttable}
\theoremstyle{definition}
\newtheorem{definition}{Definition}[section]
\titleformat*{\section}{\large\bfseries}
\titleformat*{\subsection}{\normalsize\bfseries}
\usepackage{amsthm,amssymb}
\usepackage{amsmath}

\usepackage{caption}

\usepackage[table]{xcolor}
\usepackage{array,ragged2e}
\newcolumntype{M}{@{}>{\columncolor{white}[0pt][0pt]}c@{}}

\usepackage{longtable}
\usepackage{makecell}
\usepackage{float}
\usepackage[font=footnotesize,labelfont=bf]{caption}
\title{Augmented Fairness: An Interpretable Model Augmenting Decision-Makers' Fairness} 
\date{}
\author{Tong Wang \\University of Iowa \\tong-wang@uiowa.edu\and Maytal Saar-Tsechansky\\University of Texas at Austin\\maytal.saar-tsechansky@mccombs.utexas.edu}

\newcommand{\abstractText}{
We propose a model-agnostic approach for mitigating the prediction bias of a  black-box decision-maker, and in particular, a human decision-maker. Our method detects in the feature space where the black-box decision-maker is biased and replaces it with a few short decision rules, acting as a ``fair surrogate''. The rule-based surrogate model is trained under two objectives, predictive performance and fairness. Our model focuses on a setting that is common in practice but distinct from other literature on fairness. We only have black-box access to the model, and only a limited set of true labels can be queried under a budget constraint. We formulate a multi-objective optimization for building a surrogate model, where we simultaneously optimize for both predictive performance and bias. To train the model, we propose a novel training algorithm that combines a nondominated sorting genetic algorithm with active learning. We test our model on public datasets where we simulate various biased "black-box" classifiers (decision-makers) and apply our approach for interpretable augmented fairness. 
   
  $\\[5pt]$
  Key words - fairness enhancement, interpretable machine learning, post-processing
}

\begin{document}


\onecolumn
  \begin{@twocolumnfalse}
    \maketitle
    \begin{abstract}
      \abstractText
      \newline
      \newline
    \end{abstract}
  \end{@twocolumnfalse}

\maketitle


\section{Introduction}
Machine learning is being rapidly integrated into practice across a broad set of domains to inform or entirely automate decision making. In recent years, concerns about bias in machine learning predictions have led to a growing interest in fair machine learning. Many methods have been proposed, which can be broadly described as pursuing one of three strategies. The first, ``pre-process'' strategy, pertains to directly reducing bias from the training data by learning a fair representation of the data, and then training a model on the learned representation (e.g., \cite{zemel2013learning}). A second, ``in-process'' strategy,  incorporates fairness in the machine-learning model's objective function, and trains models that directly optimize it, along with other objectives such as predictive performance \cite{zafar2017fairness}. Finally, a third ``post-process'' strategy manipulates the outputs of a model to reduce bias. 

However, in practice, a vast array of high-stake decisions, such as medical diagnoses and court decisions, are performed and ultimately vetted by human decision-makers. 
Meanwhile, prior work that primarily focuses on dealing with biased machine learning models, does not apply to mitigate bias of human decision-makers for the following constraints. First, unlike machine learning models, human decision-makers ' mapping from input to output cannot be altered or retrained, which rules out the pre-process and in-process methods. Second, for post-process methods, many existing methods require \emph{probabilistic} decisions as their input  \cite{kamiran2012data, pleiss2017fairness}; a human decision-maker, however, does not typically produce probabilities towards decisions, but rather produce discrete decisions, such as a diagnosis or a course of treatment. In addition, some existing post-processing methods require on-demand access to the model \cite{kim2019multiaccuracy}, using it as an oracle to provide real-time decisions for some constructed decisions instances during training. Human decision-makers, however, are too costly or infeasible to be actively involved during training, nor is it reasonable for human experts to produce genuine decisions for constructed (made-up) variations of decision instances. In addition, most work on fairness assumes ground truth is available for all historical data. However, ground truth for high-stake human decisions are often unavailable and expensive to acquire (e.g., via a panel of costly experts). 
Finally and most importantly, none of the existing post-processing methods provides an interpretable solution for mitigating bias. To work with human decision-makers, however, interpretability is essential since the model needs to be understandable for human experts to vet and approve. In fact, interpretability has become a de-facto requirement in practice for ML-based decision making, particularly in high-stake domains \cite{rudin2019stop}.

To summarize, in order to work mitigate the bias of human decision-makers, a model needs to 1)  not rely on on-demand access of the decision-maker and not alter the decision-maker, 2) work with non-probabilistic outputs, 3) work a small set of true labels, and 4) be interpretable.

Thus, in this paper, we propose a novel approach to mitigate human bias while satisfying all constraints above. Our approach augments (a potentially biased) human decision-maker by replacing a subset of his decisions with those of an interpretable surrogate model constructed from association rules. The interpretable model, that knows when to step in and how to make decisions, is trained on the human decision-maker's past decisions and a small subset of true labels. Thus we create a human-machine hybrid that yields a superior trade-off between accuracy and bias. We call our model the Augmented Fairness model (\textbf{AuFair}).
Note that our setting also corresponds to cases where a black-box model is used as a service, such as a proprietary commercial image recognition system \cite{kim2019multiaccuracy}.  


Formally, our setting includes a black-box decision-maker $h$, and a set of decision data $\mathcal{D} = \{\mathbf{x}_i, h(\mathbf{x}_i)\}_{i=1}^n$ where $\mathbf{x}_i$ is a vector of features for instance $i$ and $h(\mathbf{x}_i)$ is the decisions label produced by $h$. Unlike the setting considered in prior work, access to the full set of true labels is not given, but true labels can be actively acquired for selective instances from $\mathcal{D}$, under a budget constraint $B$. We propose a method that replaces a subset of classifier $h$'s decisions with decisions of an interpretable surrogate model, $g$. The collaboration of the two models aims to satisfy two key performance objectives:   \emph{predictive} generalization performance, and  \emph{fairness}. 

Towards that, we devise a multi-objective optimization algorithm for training the model. The training method starts with training data with only the labels provided by the biased decision-maker $h$, but no true labels. With budget $B$, our training algorithm iteratively acquires the labels of a small set of true labels, until the budget is exhausted. The algorithm returns a set of models on the Pareto frontier of predictive performance versus fairness.

Our contributions are as follows. We develop a novel method for mitigating the bias of an unfair decision-maker, that is inspired by the goal to effectively augment human decision-makers in high stake decision domains. This new setting gives rise to several key properties of our model, which are distinct from those of prior works. An AuFair model is interpretable, only needs predicted labels from a decision-maker at the outset, and is trained on a limited set of true labels.
To the best of our knowledge, this is the first work to propose and develop a method for such human decision-maker augmentation. In addition, we provide empirical evaluation of our approach, and demonstrate its efficacy in producing superior fairness and accuracy tradeofs for different problem domains. 


\section{Related Work}\label{sec:ref}
The fast and extensive integration of machine learning in a wide variety of domains has lead to a heightened focus on fairness in decision making fairness. Previous research has focused primarily on combating the biases of machine learning models, caused by bias in the training data or from the machine learning algorithm.  Interested readers can refer to \cite{mehrabi2019survey,barocas-hardt-narayanan} for more comprehensive review.

Various definitions of fairness have been proposed for different contexts. For example, \emph{demographic parity} means that the proportion of each segment of a protected feature (e.g., gender) should receive the favored outcome at equal rates. Parity in both false negative rate and false positive rate is referred to as \emph{equal odds} \cite{lohia2019bias}. Its relaxation, equality in only true positive rates, is referred to as \emph{equal opportunity}, which describes that subjects in different segments of a protected feature have an equal probability of being correctly classified as favorable. While our framework applies with both demographic parity and equal odds, in this paper, we mainly focus on the equal opportunity definition and define unfairness as the difference in the true positive rates across different groups defined by a protected feature.

To achieve fairness, existing works fall into three categories, pre-processing \cite{zemel2013learning, calmon2017optimized, madras2018learning,wangrepairing}, in-process mechanisms \cite{berk2017convex}, and post-processing \cite{hardt2016equality,lohia2019bias}.
Ours is also a post-processing method. 
One of the earliest works on post-processing methods is rejection option-based classification (ROC) \cite{kamiran2012data}. This approach assumes that the discrimination occurs when a model is least certain of a prediction, i.e., around the decision boundary (classification threshold). Therefore, ROC modifies predictions when the input is near the decision boundary: the favorable outcome is given to the privileged group, or the unfavorable outcome is given to the unprivileged. However, this approach requires probabilistic outputs from the model, which are often not available, including in particular when the decision-maker is human. Other works from this category similarly require probabilistic outputs from a model for post-hoc calibration, such as \cite{pleiss2017fairness}.
Another well-known work \cite{hardt2016equality} proposes a more straightforward way to achieve fairness by randomizing a fraction of individuals in the favorable and unfavorable group to obtain the desired true positive rates and false positive rates. 
Recent works propose methods that consider fine-grained, group-level fairness, and often assumes access to the black-box model, which means the unfair decision-maker must be involved in the training process. For example,
In \cite{noriega2019active}, a decision-maker adaptively acquires information according to the needs of different groups or individuals towards balancing disparities in classification performances. This approach substantially outperforms randomization based classifiers such as the model in \cite{hardt2016equality}. Another such approach by
\cite{kim2019multiaccuracy} proposes a framework for auditing and post-processing prediction models for multiaccuracy. However, the methods mentioned above cannot be used on human decision-makers since they either require full access to the true labels, or work with probabilistic outputs, or need to actively involve the biased model in training.

 Among the existing methods,  \cite{hardt2016equality} is closest to the setting we describe: it can work with predicted labels and does not require to involve the biased model in the training. However, this method does require full access to true labels. We will later experiment with subsets of data as input to compare with our AuFair model. 
\section{Augmented Fairness}
We are provided with a decision-maker $h: \mathbb{R}^p \rightarrow \{0,1\}$ that may be biased in the protected feature $z$. We consider binary classification in this paper and let 1 be the favored outcome for the subjects on which we aim to improve fairness. We do not have access to model $h$'s mapping, nor can this mapping be revised. However, we have access to a set of historical decision data for which the labels produced by $h$ are provided, $\mathcal{D} =  \{\mathbf{x}_i, h(\mathbf{x}_i)\}_i^n$,  where $\mathbf{x}_i \in \mathbbm{R}^p$ is a vector of $p$ features, and $h(\mathbf{x}_i)$ is the output of model $h$. 
 To mimic the realistic constraint, the true labels for $\mathbf{x} \in \mathcal{D}$ are not provided at the outset, but can be acquired at a cost, for a given total budget of $B$. We assume the unit cost in this paper, so that one can acquire the true labels of up to $B$ instances. Formally, let $Q \subset \{1,\cdots, n\}$ be the indices of data points whose true labels, $\mathbf{y}_Q: = \{y_i\}_{i\in Q}$, are acquired by the training algorithm and $|Q| = B$.  The complete query algorithm is outlined in Section \ref{sec:main}.

Our goal is to produce an augmented fairness decision-maker $f$ that \textbf{augments} $h$ by partially substituting $h$, for some input data, with a model $g: \mathbb{R}^p \rightarrow \{0,1\} $ that we construct, such that the predictions provided by either $h$ or $g$ on the entire input space achieve a superior trade-off between predictive performance and fairness. In addition, we require $g$ to be interpretable so that its reasoning can be easily understood and vetted by human decision-makers. 
In this paper, we achieve these goals by constructing $g$ as a rule-based model, where each rule is a conjunction of conditions. The decision-making process follows that of hybrid models \cite{wang2019hybrid,wang2019gaining}. Specifically, we construct two rule sets towards $g$, a \emph{positive rule set}, $\mathcal{R}_+$, and a \emph{negative rule set}, $\mathcal{R}_-$.  If $\mathbf{x}_i$  satisfies any rules in $\mathcal{R}_+$, it is classified as positive (favorable class). Otherwise, if it satisfies any  rules in $\mathcal{R}_-$, it is classified as negative (unfavorable class). A decision on $\mathbf{x}_i$ produced from rules in $g$ is denoted as $\widehat{y_g}_i$. If $\mathbf{x}_i$ does not satisfy any rules in $\mathcal{R}_+$ or $\mathcal{R}_-$ (and thus $g$ decides to let $h$ decide) then $\mathbf{x}_i$ is sent to the original decision-maker $h$ to generate a decision $h(\mathbf{x}_i)$. 
The predictive process of the hybrid model $f$ is outlined below.
\begin{align}
\text{Augmented decision-maker } f: \quad&\textbf{if } \mathbf{x}_i \text{ obeys } \mathcal{R}_+, Y = 1 \notag \\
&\textbf{else if }\mathbf{x}_i \text{ obeys } \mathcal{R}_-, Y = 0 \notag \\
&\textbf{else }Y= h(\mathbf{x}_i)
\end{align}
Table \ref{tab:example} shows examples of an AuFair model learned from the Adult dataset.
\begin{table}[h]
\caption{The protected feature is sex and the favorable outcome is income $\geq$ \$50K. For the Aufair $f$, The  error is 0.17 and the bias is 0.04, evaluated on the test set. For the original model $h$, the error is 0.16 and the bias is 0.18 on the test set. }\label{tab:example}
\small
\begin{tabular}{llc}
\toprule
        & \multicolumn{1}{c}{\textbf{AuFair Model}}                                                                    \\\hline
\textbf{If} &  degree $\neq$ some college \textbf{\emph{and}} number of years of education $\geq 10$ \textbf{\emph{and}}   relationship = Wife \\
& \textbf{\emph{OR}} degree = PhD \textbf{\emph{and}} number of years of education $\geq 13$ \\
        & $\rightarrow$ \textcolor{black}{$Y = 1$} (income $\geq$ \$50K)  \\ 
\textbf{Else if} & degree $=$ high school \textbf{\emph{and}} marital status = divorced  \\
        & $\rightarrow$ \textcolor{black}{$Y = 0$} (income $<$ \$50K)  \\ 
\textbf{Else}    &\textcolor{black}{$Y = h(\mathbf{x})$}   \\ \bottomrule                
\end{tabular}
\end{table}
\paragraph{Multi-objective Augmented Fairness Formulation}
We aim to optimize two objectives when constructing $\mathcal{R}_+$ and $\mathcal{R}_-$:  predictive performance and  fairness. Note that both metrics can be customized with existing or task-specific definitions. In this paper, we consider predictive error as a measure of predictive performance, and (un)equal opportunity as the (bias) fairness measure, defined as follows:
\begin{align}
&\textbf{Error}:\quad    \mathcal{L}(f; \mathcal{D}, \mathbf{y}_Q) = \sum_{i\in Q} \mathbbm{1}( y_i \neq f(\mathbf{x}_i)),\\
&\textbf{Bias}:\quad \Phi(f; \mathcal{D}, \mathbf{y}_Q) = |  Pr(\hat{Y} = 1|z = 0, Y = 1) - Pr(\hat{Y} = 1| z = 1, Y = 1)|
\end{align}
For convenience, henceforth we will ignore the dependence on $\mathcal{D}, \mathbf{y}_Q$ in the notation.
To simultaneously optimize for the two objectives, we formulate a multi-objective optimization. Multi-objective optimization deals with problems in which conflicting objectives ought to be simultaneously maximized or minimized, and their optimum is composed by a set of tradeoff solutions known as Pareto optimal set. 
Our multi-objective optimization can be simply stated as:
\begin{equation*}
  \min  \Phi( f),  \mathcal{L}(f) \quad\quad (M)
\end{equation*}
Our goal is to obtain a set of Pareto optimal set of solutions, denoted by $\mathcal{F}^* = \{f_1, \cdots, f_{|G|}\}$, to program M, such that $\forall f \in \mathcal{F}^*, \nexists f^\prime \in \mathcal{F}^*$ that \emph{dominates} $f$. The notion of ``dominance'' is defined below.
\begin{definition} An AuFair $f$ is dominated by another AuFair model $f^\prime$, i.e., $f^\prime \prec f$, if $\mathcal{L}(f^\prime)< \mathcal{L}(f)$ and $\Phi(f^\prime)\leq\phi(f)$ or if $\mathcal{L}(f^\prime)\leq \mathcal{L}(f)$ and $\Phi(f^\prime)<\phi(f)$.
\end{definition}
Note that the solution $\mathcal{F}^*$ is a set of models instead of one optimal model. Models in $F^*$ cover a large range of error and fairness tradeoffs for users to select from, to satisfy any task-specific preferred tradeoff between the two objectives.

\section{Multi-objective Optimization with Active Learning}
An effective method to solve complex multi-objective optimization problems corresponds to non-exact algorithms, i.e.,  metaheuristics \cite{blum2003metaheuristics}, that include evolutionary algorithms and swarm intelligence methods (such as ant colony optimization, or particle swarm optimization).
However, our model cannot directly use any of the existing algorithms because, at the outset, we begin with an empty set of true labels.  Thus, our approach also incorporates a selective true label acquisition policy into the training algorithm, acquiring the true labels of a particularly informative subset of instances from $\mathcal{D}$.

Specifically, we design a training algorithm based on the main steps of a well known genetic algorithm framework - Nondominated Sorting Genetic Algorithm II (NSGA-II)- along with a query algorithm that iteratively acquires the labels of a small set of instances, until the budget is exhausted.  We call the algorithm \emph{Active Nondominated  Sorting Genetic Algorithm -- $a$-NSGA}. Our algorithm returns a set of non-dominated AuFair models, $\mathcal{F}^*$.

\vspace{-2mm}\subsection{Pre-Steps for the Algorithm}
  We first induce two sets of candidate rules from $\mathcal{D}$: $\Upsilon_+$ for positive instances and $\Upsilon_-$ for negative instances. $\Upsilon_+$ and $\Upsilon_-$ contain rules with high precision and our algorithm selects rules from these sets to form $g$ in each solution. To induce $\Upsilon_+$ and $\Upsilon_-$ we use an off-the-shelf rule mining algorithm, FPGrowth, which returns frequently co-occurring conditions from $\mathcal{D}$, satisfying a minimum support and maximum length constraint. Other rule mining algorithms can also be used. Following prior work using this algorithm, in this paper we set the maximum length to 3 and the minimum support to 5\%  \cite{wang2017bayesian, wang2019hybrid}. 

In addition to $\mathcal{D}$, the algorithm also requires one to specify the budget $B$. While the budget constraint is typically given, we explored how different budgets $B$ affect the algorithm performance. As expected, the general observation is that larger $B$ guarantees better performance. Another parameter, $\tau$ refers to the number of iterations until the algorithm makes new acquisitions of true labels. In the experiments reported here we set $\tau$ to 2. 

\vspace{-2mm}\subsection{Main Loop}\label{sec:main}
As discussed, our approach uses genetic algorithm to iteratively select elite parents, namely advantageous rule populations, to produce offsprings, such that the population increasingly improves over subsequent iterations. Simultaneously, selected true labels are queried progressively to improve the evaluation of each solution and the selection of elite solutions towards reproduction. Ultimately, the non-dominated solutions -- i.e., the solutions that are not worse in both error and fairness than any other solutions--  are returned as the outputs of the algorithm.
Specifically, the elements of the genetic algorithm in our setting are:
\begin{itemize}
    \item \textbf{Solution}: an AuFair $f$ consists of $g$ and $h$. $g$ is comprised of a positive and a negative rule sets. In this algorithm, only $g$ is iteratively updated and learned from data.
    \item \textbf{Population}: the population is a set of AuFair $f$, such that each solution includes the same(given) $h$ and different rule sets $g$.
    \item \textbf{Fitness Comparison}: all solutions in the population are evaluated for error and unfairness, and each pair of solutions can be ranked based on  Definition 3.1.
\end{itemize}

Let $P_t$ denote the set of parent solutions, and $O_t$ is the offsprings produced from solutions in $P_t$ (using produce-offsprings function which we will describe later). 
 $a$-NSGA initializes $P_0$ with a set of randomly generated solutions and derive $O_0$ from $P_0$. Then in each iteration,  $P_t$ and $O_t$ are combined into a set $R_t$, which includes all solutions that will be processed in the current iteration. Only $N$ solutions can reproduce, and hence we first sort $R_t$  using a nondominated sorting algorithm that sorts the solutions according to Definition 3.1. The purpose of the sorting algorithm is to divide the solutions to different frontiers $F = (F_1, F_2, \cdots)$ in terms of the fairness and accuracy performances. Those that are not dominated by $k$ solution are on frontier $k$, $k\geq 1$.   Solutions on frontiers that are ranked higher have better quality and should be prioritized in reproducing new offsprings. For those that are on the same frontier, another sorting algorithm is used that computes the crowding distance for each solution, considering its average distance to the nearby points along each of the objectives -- fairness and accuracy. This principle aims to identify the degree to which other solutions exist in the population that yield similar outcomes along the frontier. Thus, solutions with larger crowding distance are ranked higher since the algorithm encourages diversity in the reproducing step. Then, solutions in $R_t$ are sorted, and the best $N$ solutions have the right to reproduce. The two sorting functions are the same as the ones proposed in \cite{meyarivan2002fast}.

 \vspace{-2mm}
\paragraph{Producing Offsprings} In genetic algorithms, the process of producing offsprings includes selection, crossover, and mutation. We adopt the core principles of these functions and customize then to our model.
Given the $N$ solutions sorted and selected from $R_t$, we repeat the following reproducing step $N$ times to produce $N$ offsprings. Each time, we randomly select two solutions, and the better one is selected to become a parent, denoted as $p_1$. We repeat the process to select the second parent, $p_2$.   The two parents then produce an offspring by a crossover followed by mutation. In the crossover function, each rule in a parent has a probability of 0.5 to be included in their offspring. Then in the mutation step, the selected rule has probability 0.9 to be kept. In addition, we randomly select $n$ rules from those not in the offspring set, where $n$ is drawn from 1 to the average length of $p_1$ and $p_2$. This reproducing function ensures that some elitism is maintained from two good parents while adding some randomization in their offspring. 

 \vspace{-2mm}
\paragraph{Query Method}
Our setting is distinct from existing genetic algorithms for solving multi-objective optimization because no true labels are available initially. We start with an empty set of labels and iteratively query $b$ labels at a time until the budget is used. Let $Q_t$ represent the indices of instances whose true labels are queried till iteration $t$ and $Q_0 = \emptyset$. Our query selection criterion considers the uncertainty in the prediction by the best current solutions, i.e., solutions in the first front $F_1$. Towards this principle, we follow the query-by-bagging algorithm proposed in  \cite{abe1998query}. Specifically, we create 10 bootstrap samples from the training data. For each solution, we obtain its probabilistic outputs for each instance, which is the probability of the rule that captures it. For example, if 80\% of the instances in a bootstraped sample captured by a rule are positive, then the rule outputs probability 0.8, and instances captured by it all have a probability estimation of 0.8.  We do that for each of the 10 datasets and obtain 10 probability assignments for each data point. An uncertainty score can then be obtained for every data point by computing the variance over the 10 probabilities. 
Then we sample $b$ data points from the unlabeled data, indexed by $\{1,\cdots, n\}\backslash Q_t$, with the sampling weight determined by the uncertainty score. 

The complete algorithm is summarized in Algorithm \ref{alg:search}.
\begin{algorithm}[h!]
\caption{Active Nondominated Sorting Genetic Algorithm ($a$-NSGA)}\label{alg:search}
\begin{algorithmic}[1]
\STATE \textbf{Input:} $\mathcal{D}$, query interval $\tau$, budget $B$, label acquisition batch size $b$
\STATE \textbf{Initialize:}  
           \STATE $\Upsilon_+, \Upsilon_- \leftarrow \text{FPGrowth}(\mathcal{D}, \text{minsupp} = 5\%)$
                        \STATE $P_0  \leftarrow$ $N$ solutions by randomly drawing 1 to 5 rules from $\Upsilon_+$ and $\Upsilon_-$, respectively
            \STATE $O_0 = $produce-offsprings($P_0$)
            \STATE $t = 0$ and $Q_0 = \emptyset$
            \WHILE{$|Q_t|< B$}
            \STATE $R_t = P_t \cup O_t$ \quad\quad\quad\quad\quad\quad\quad\quad\quad $\triangleright$ combine parent and offspring population
            \STATE $F = $fast-non-dominated-sort$(R_t)$\quad$\triangleright F = (F_1, F_2, \cdots)$, all nondominated fronts of $R_t$ 
            \IF {$t \geq \tau$}
            \STATE$ t= 0 $\quad\quad\quad\quad\quad\quad\quad\quad\quad\quad\quad\quad$\triangleright$ reset the counter 
            \STATE $Q_t = Q_t \cup $query-labels$(F_1, b, Q_t)$
            \ENDIF
            \STATE $P_{t+1} = \emptyset$ and $i = 1$
            \WHILE {$|P_{t+1}|+|F_i| \leq N$}
            \STATE crowd-distance-assignment($F_i$) \quad $\triangleright$  calculate crowding-distance in $F_i$
            \STATE $P_{t+1} = P_{t+1} \cup F_i$ \quad\quad $\triangleright$ include $i$-th nondominated front in the parent population
            \STATE $i = i+1$ \quad\quad\quad\quad\quad $\triangleright$ check the next front for inclusion
            \ENDWHILE \quad\quad\quad\quad\quad\quad $\triangleright$ until the parent population is filled
            \STATE Sort($F_i$) \quad\quad\quad\quad\quad $\triangleright$ sort solutions based on the crowded-comparison
            \STATE $P_{t+1} = P_{t+1} \cup F_i[1:(N - |P_{t+1}|)]$ \quad\quad $\triangleright$ choose the first $(N-|P_{t+1}|)$ elements of $F_i$
            \STATE $O_{t+1} =$ produce-offsprings($P_{t+1}$)  \quad\quad $\triangleright$ use selection, crossover and mutation to create offspring population $Q_{t+1}$
            \STATE $t = t+1$
            \ENDWHILE
    \STATE \textbf{Output:} $\mathcal{R}^*$
\end{algorithmic}
\end{algorithm}
\normalsize
  \section{Empirical Evaluations}
 We evaluate our proposed model on public datasets and compare its performance to that of baseline alternatives.
 We use two publicly available datasets, the Adult data \cite{kohavi1996scaling} and Recidivism data \cite{langan1992recidivism}. For the adult dataset, the target variable is whether the person's income is at least \$50K per year. We create two problems considering either race or sex as the protected feature. For the recidivism dataset, the target variable is whether an individual will re-offend, and we consider the sex as the protected feature. We thus create three predictive tasks, each corresponding to a domain and a protected feature. For each task, we conduct 5-fold cross-validation, and we further split each training fold to 80\% training data and 20\% validation data.
 
 For a given training set $\mathcal{D} = \{\mathbf{x}_i, y_i\}_{i=1}^n$, we first train a Lasso model $h$ to represent a decision-maker, which produces a set of predicted labels (decisions) $\{h({\mathbf{x}_i})\}_{i=1}^n$. We then apply our approach to produce an AuFair model $f$. Specifically, we use the training set to obtain a set of solutions (Pareto frontier on the training set), and then evaluate them on the validation set so as to choose the Pareto frontier. The selected Pareto frontier is subsequently applied to augment the decision-maker $h$ and make inferences over the test. 
We evaluate our approach relative to several alternatives. The first baseline was proposed by \cite{hardt2016equality}, and it forms a linear program to solve for four decision variables. The four decision variables are true and false positive rates in both the protected and unprotected group. The approach then randomly flips the labels produced by $h$ to match the four probabilities. We refer to this method as \textbf{EOP} and use the implementation from \cite{aif360-oct-2018}. Note that unlike our method, which is flexible to tradeoff error and bias, \textbf{EOP} aims to achieve absolute equal opportunity, thus producing a single solution for each dataset. Two additional baselines refer to simply flipping the protected feature (to alleviate bias) before forwarding the instance to the decision-maker $h$. $\tilde{h}_1$ denotes the results for when all protected features are set to 1, and $\tilde{h}_0$ is when  all protected features are set to 0. 

\vspace{-2mm}\subsection{Error - Bias Tradeoff}
\begin{figure}[b]
  \centering
  \includegraphics[trim = 0 1.2cm 0 0.5cm, clip, width=\textwidth]{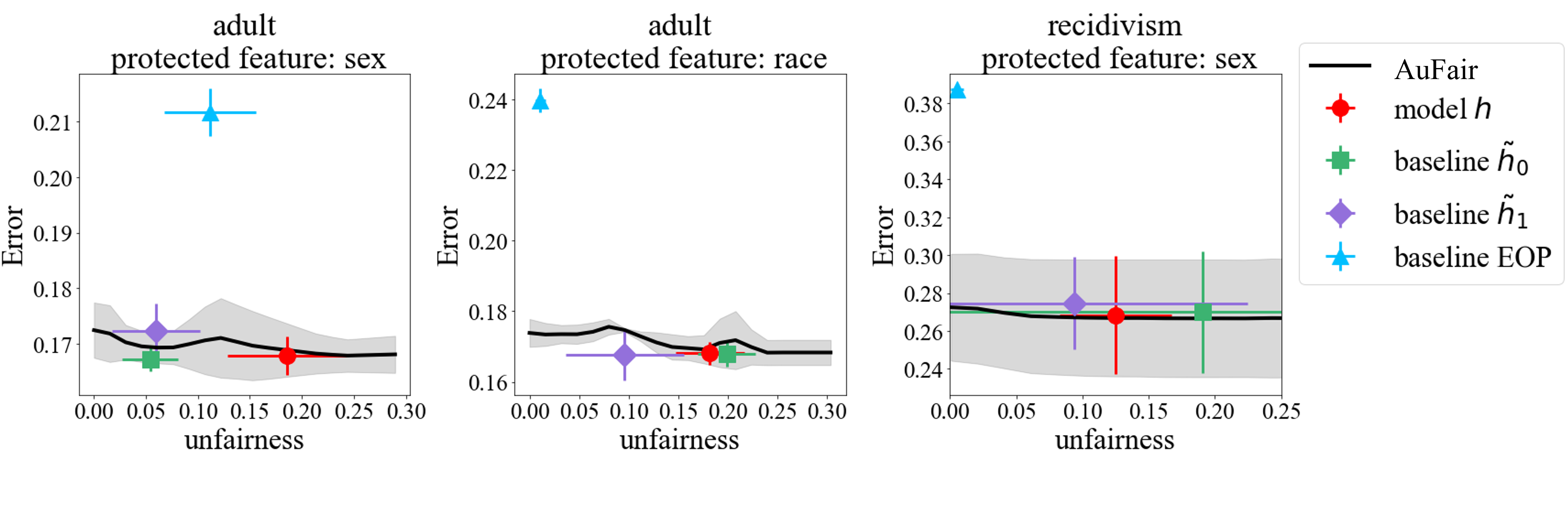}
  \caption{The Pareto frontier obtained with different percentage of true labels.}
  \label{fig:tradeoff}
\end{figure}

 Recall, that in this work bias is measured by the absolute difference between the true positive rate of the protected and unprotected group. Each baseline method returns a model which we can evaluate its bias and error on the test set. For AuFair method, the algorithm returns a set of AuFair models. We thus obtain a curve that characterize the tradeoff  between bias and error.
Figure \ref{fig:tradeoff} shows the mean and standard deviation  evaluated on 5-folds for all models. Because the alternatives assume labels for 100\% of the data, we first evaluated our approach with 100\% true labels, and later consider different label acquisition budgets. 
 
Figure \ref{fig:tradeoff} show that AuFair yields significantly better tradeoffs than enabled by the EOP  baseline. Compared to other baselines, AuFair achieves similar error but can reach significantly lower bias. Recall that we derive $h$ based on a model that optimizes predictive performance. Hence, as compared to the baselines that flip the protected features before $h$ infers the outcome, our results show that our method offers important flexibility in providing an array of solutions with superior fairness. Specifically, while the $\tilde{h}_0$ and $\tilde{h}_1$ offer a single level of fairness and accuracy, our approach offers multiple solutions with either no unfairness, as well as with significantly lower unfairness than possible with the alternatives, and which may be advantageous to satisfy different tasks-specific preferences.

\vspace{-2mm}\subsection{Performance under a Budget Constraint}
\begin{figure}[h]  
  \centering
  \includegraphics[trim = 0 1.5cm 0 0.5cm, clip, width=\textwidth]{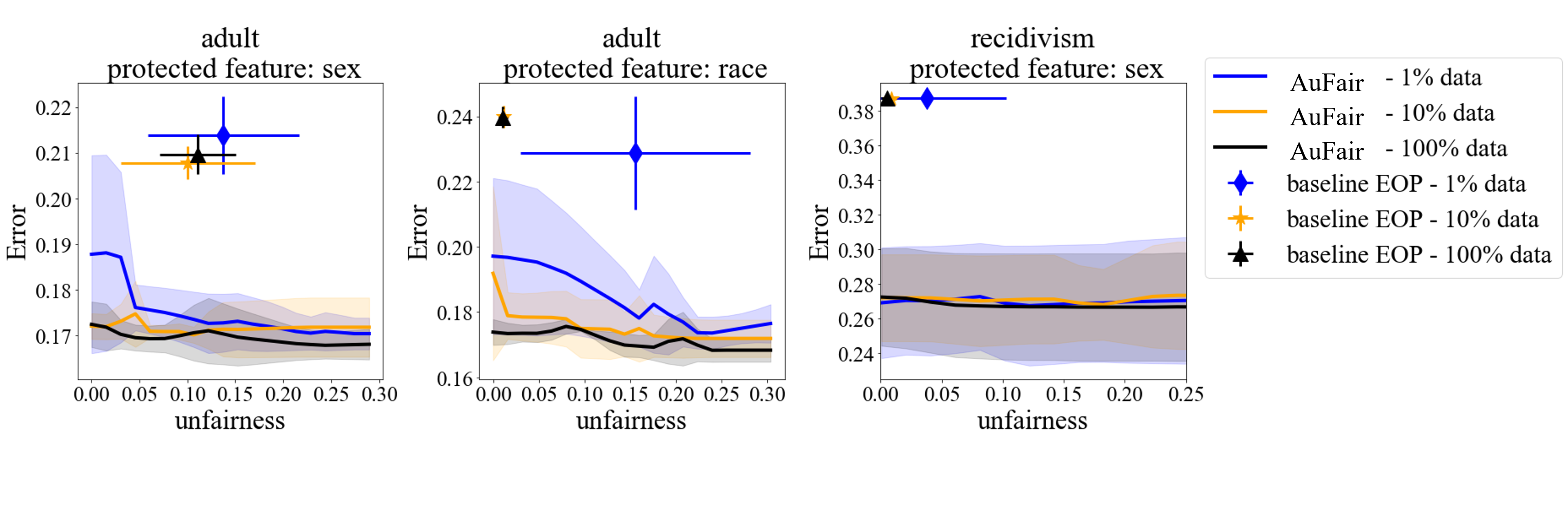}
  \caption{The Pareto frontier obtained with different percentage of true labels.}
  \label{fig:query}
\end{figure}

We evaluated our approach under different budget constraints, where the available budget allows to query 1\%, 10\%, and 100\% of the true labels. We apply the same amounts of correct labels to the baseline EOP. Our results are As shown in  Figure \ref{fig:query}.

As shown, AuFair's advantage relative to the alternatives under limited ground truth labels is strong, and AuFair produces solutions that offer either both lower error and bias, or otherwise offer significantly lower error. In addition, the results show that with more true labels, AuFair is able to ``push'' the Pareto curve down, particularly between acquiring 1\% and  10\% of the true labels. The variance is similarly decreased with more true labels. The label acquisition policy appears to acquire particularly informative labels given acquiring more than 10\% of the true labels, does not yield significant benefits. A similar phenomenon is observed for the baseline method. Furthermore, our results show that even if we only query 1\% of true labels, AuFair's performance is satisfying as compared to when baselines use 100\% of the true labels. Our proposed AuFair approach entirely dominates the EOP baseline under all budget constraints.


\section{Conclusion}
We introduce a novel approach for Augmented Fairness: augmenting a biased, ``black-box", human decision-maker with interpretable decision rules, so as to achieve advantageous fairness and accuracy tradeoffs. Our problem setting differs meaningfully from those considered in prior work:  the decision-maker cannot be made available on-demand nor altered, the true labels of historical data are unavailable, only a subset can be acquired, and the augmented inferences must be easily understandable by humans. This setting, which corresponds to when a human is the decision-maker, is one which, to our knowledge, none of the existing fairness methods have considered. 
Our proposed AuFair method learns a set of interpretable decision rules that substitute the black-box decision-maker on some test instances, while providing understandable predictions. We devise a multi-objective training algorithm that returns a set of solutions of different tradeoffs between fairness and error. Experimental results show that our model achieves meaningfully improved performance over the state-of-the-art alternatives, along with important flexibility for identifying task-specific, advantageous tradeoffs. 
\bibliographystyle{plain}

\end{document}